\documentclass[12pt,a4paper]{article}
\usepackage{arxiv}
\usepackage[T1]{fontenc}
\usepackage{graphicx}
\usepackage{amsmath}
\usepackage{setspace}
\usepackage{hyperref}
\hypersetup{
	colorlinks=true,
	linkcolor=blue,
	citecolor=blue,
	urlcolor=blue
}

\begin{document}
\title{\textbf{The Slop Paradox: How Synthetic Standardization Erodes Clinical Uncertainty and Cross-Modal Alignment in AI-Rewritten Radiology Reports}}
%

\author{
	Samar Ansari\\
	School of Computing and Engineering Sciences\\
	University of Chester\\
	Chester, CH1 4BJ, United Kingdom \\
	\texttt{m.ansari@chester.ac.uk}
}
%
%
%
%
\maketitle              
\begin{abstract}
AI-assisted clinical documentation tools increasingly summarize, standardize, and reformat radiology reports using large language models (LLMs). We present a controlled measurement of the resulting information degradation. Using 450 chest X-ray reports from the Indiana University dataset, we generate synthetic versions via three realistic LLM rewriting tasks: EHR summarization, standardized rewriting, and teaching case preparation. We measure entity erosion (via medical NER), hedging collapse (loss of clinical uncertainty language), and cross-modal alignment degradation (via BiomedCLIP image-text similarity). Our central finding is a dissociation between information loss and cross-modal fidelity. EHR summarization is the most destructive at the content level, eroding 51.4\% of clinical entities and 43.7\% of hedging language, yet it preserves image-text alignment almost entirely (a 2.5\% drop). The two tasks meant to produce cleaner training data, standardized rewriting and teaching case preparation, do the reverse: they preserve more entities (26.8\% and 29.3\% eroded) but cause 14.9--16.5\% alignment drops, six to seven times those of EHR summarization. We term this the \emph{slop paradox}: rewriting that makes clinical text look cleaner for multimodal training is precisely what pulls it away from the image. Contrary to our pre-specified hypothesis, rare pathologies were not preferentially degraded: across nine rare-versus-common comparisons, no difference survived multiple-comparison correction, and nominal differences ran in the opposite direction (common $>$ rare), so contamination is invisible to condition-specific monitoring. The dominant determinant of degradation is the type of AI rewriting task, not the clinical content. These findings bear on multimodal medical AI dataset construction and the governance of AI-assisted clinical documentation.

\end{abstract}
%
%
%

\section{Introduction}\label{sec:introduction}

Large language models (LLMs) are increasingly integrated into clinical workflows. AI-assisted documentation tools summarize radiology reports for electronic health records (EHRs), rewrite clinical notes in standardized formats, and generate teaching materials from real patient cases~\cite{de2025improving,alkhalaf2024applying,tanno2025collaboration}. These applications respond to genuine pressure: an estimated 97\% of UK imaging departments cannot keep pace with reporting workloads~\cite{sloan2024automated}. AI digital scribes built on GPT-4o have already generated tens of thousands of clinical documents in deployed systems~\cite{de2025improving}, and Llama-2-based summarization with retrieval augmentation reaches 99\% accuracy on structured EHR fields~\cite{alkhalaf2024applying}.

A critical question, however, remains underexamined: what happens to the clinical information content of a radiology report when it passes through an LLM? Goodman et al.~\cite{goodman2024ai} warned that AI-generated clinical summaries introduce risks beyond simple inaccuracy, including ``complete-the-narrative'' errors where models fabricate clinically plausible but unsupported details. The broader phenomenon of AI-generated content degrading information quality has been documented across domains under the framework of ``AI slop'' and recursive data contamination~\cite{resnik2026vicious,anon_slop2025}, with evidence that as few as 250 corrupted documents can introduce persistent biases into large-scale models. In the medical domain, the consequences of such degradation are not cultural homogenization but potential diagnostic failures.

Despite growing concern, no study has systematically measured the specific types and magnitude of information loss that occur when clinical radiology reports are synthetically rewritten by LLMs. Prior work on automated radiology report generation has focused on generation quality metrics such as BLEU and clinical accuracy scores~\cite{sloan2024automated,goswami2025medivlm,nakaura2024preliminary}, rather than on the information that is \emph{destroyed} in the process. Work on clinical uncertainty language has established that hedging expressions appear in approximately 35\% of radiology reports~\cite{sloan2024automated} and carry specific, rankable diagnostic significance~\cite{rabaey2025modeling}, but no study has examined whether AI rewriting preserves or eliminates this language. In the multimodal setting, models such as BiomedCLIP~\cite{zhang2023biomedclip} rely on paired image-text data, yet no study has quantified how synthetic report text degrades the alignment between clinical images and their paired reports.

This paper addresses these gaps through a controlled experiment on the Indiana University Chest X-Ray dataset~\cite{demner2016preparing}. We generate synthetic versions of 450 radiology reports using three LLM-based rewriting tasks that simulate realistic contamination vectors: EHR summarization, standardized clinical reformatting, and teaching case preparation. We measure three dimensions of information degradation: \emph{entity erosion} (clinical entities lost), \emph{hedging collapse} (clinical uncertainty language lost), and \emph{cross-modal misalignment} (degradation of image-text alignment via BiomedCLIP cosine similarity). We additionally test whether these effects disproportionately affect rare versus common pathologies, a hypothesis motivated by distributional narrowing observed in model collapse research~\cite{ott2026context,anon_slop2025}.

Our central finding is a dissociation between how much information a rewriting task destroys and how much it degrades image-text correspondence. EHR summarization is the most destructive task at the level of content, eroding 51.4\% of clinical entities and 43.7\% of hedging language, yet it preserves cross-modal alignment almost entirely (a 2.5\% drop), because it retains the core diagnostic impression. The two tasks explicitly intended to produce cleaner data for downstream model training, standardized rewriting and teaching case preparation, behave in the opposite way: they preserve more clinical entities but, by expanding and restructuring the report, pull the text 14.9--16.5\% further from the paired image, six to seven times the drift caused by EHR summarization. We term this the \emph{slop paradox}: the rewriting that makes clinical text look cleaner for multimodal training is precisely what degrades the image-text correspondence that such training depends on. We further find that this degradation does not preferentially target rare pathologies; no rare-versus-common difference survived multiple-comparison correction (Section~\ref{sec:results}), making contamination invisible to condition-specific performance monitoring. These results have direct implications for multimodal medical AI training and for the governance of AI-assisted clinical documentation. 

\section{Related Work}\label{sec:related}

\textbf{Automated radiology report generation} has advanced with transformer-based architectures and vision-language models. Sloan et al.~\cite{sloan2024automated} provide a comprehensive review noting that the field still lacks domain-specific evaluation metrics for clinical validity. Tanno et al.~\cite{tanno2025collaboration} demonstrate that vision-language models can produce reports that experts frequently rate as comparable to human-written ones. In a small comparative study ($n=28$), Nakaura et al.~\cite{nakaura2024preliminary} found that GPT-4 achieved only 54\% Top-1 diagnostic accuracy on radiology reports versus 100\% for human radiologists. Goswami et al.~\cite{goswami2025medivlm} introduced MediVLM, demonstrating that synthetic clinical text can serve as surrogate supervision for training downstream models. These studies focus on generation quality rather than measuring information loss during the generation process.

\textbf{Clinical uncertainty and hedging language} is established as a deliberate communicative practice in radiology, with uncertainty expressions present in approximately 35\% of all reports~\cite{sloan2024automated}. Rabaey et al.~\cite{rabaey2025modeling} developed the Lunguage++ framework to quantify explicit uncertainty through LLM-based rankings of 42 hedging phrases, showing they convey specific probability levels (mean 0.459 for tentative findings). Despite this established importance, no prior study has measured whether AI rewriting preserves or destroys hedging language.

\textbf{Synthetic data and model collapse} risks are documented across AI domains. Ott~\cite{ott2026context} formalizes the connection between in-context learning and model collapse, showing that synthetic data induces distributional narrowing. Sizikova et al.~\cite{sizikova2024synthetic} survey synthetic data in radiological imaging, presenting it as a solution to data scarcity while acknowledging open questions about fidelity. A related study on AI-generated citation fabrication~\cite{anon_cite2026} demonstrated that 100\% of fabricated citations evade expert peer review through compound failure modes, illustrating the same contamination dynamic in scientific knowledge infrastructure. Prior work on AI-generated content pollution~\cite{anon_slop2025} established a framework for recursive contamination, demonstrating feedback loops that degrade information quality across domains. The present study extends these concerns to the clinical domain by empirically measuring a specific failure mode: information loss at the point of synthetic text generation, before any recursive training occurs.

\textbf{Cross-modal alignment in medical AI} has been studied through models like BiomedCLIP~\cite{zhang2023biomedclip}, pretrained on 15 million biomedical image-text pairs. Li et al.~\cite{li2026tgialign} proposed TGIAlign for cross-modal semantic alignment in medical vision-language tasks. Pandey et al.~\cite{pandey2026biomedclip} showed that fine-tuning BiomedCLIP improves cross-modal integration for diagnostic tasks. These studies optimize alignment assuming high-quality text inputs; none examines how synthetically degraded text affects alignment. Our work addresses this gap by measuring alignment drops caused by different types of synthetic rewriting.

\section{Methodology}\label{sec:methodology}

\subsection{Dataset and Stratification}
We use the Indiana University Chest X-Ray dataset~\cite{demner2016preparing} (3,826 reports with paired frontal chest X-ray images and MeSH annotations). MeSH terms are hierarchically structured (e.g., ``airspace disease/lung/base/right/mild''); we extracted root pathology terms by removing anatomical and severity qualifiers. Root pathologies appearing in $\geq$5\% of reports were classified as \emph{common} (8 conditions) and those in $<$2\% as \emph{rare} (323 conditions); root labels in the intermediate 2--5\% band (e.g., pleural effusion, pulmonary nodule, airspace disease) were treated as neither. Each report was then assigned to one group by the following rule: any report carrying at least one rare label was assigned to the \emph{rare} group (rare taking precedence even when common labels co-occurred); a report with a common label but no rare label formed the \emph{common} group; and a report with neither a rare nor a common label formed the \emph{normal} group. The normal group is dominated by no-finding studies but also contains reports whose only labels fall in the 2--5\% mid-frequency band. From the full dataset we drew a stratified random sample of 450 reports (150 per group; random seed 42), providing power $>$0.80 at $\alpha=0.05$ (two-sided) for medium effects ($d \geq 0.4$) in Mann-Whitney $U$ tests. We note in Section~\ref{sec:discussion} that the precedence rule and the heterogeneity of the normal group attenuate the rare-versus-common contrast.

\subsection{Synthetic Report Generation}
We generated synthetic versions using Gemini 2.5 Flash with three prompts simulating realistic contamination vectors:

\begin{itemize}
    \item \textbf{EHR Summary}: ``Summarize the following radiology report for inclusion in the patient's medical record.'' Simulates AI-assisted clinical documentation.
    \item \textbf{Standardized Rewrite}: ``Rewrite the following radiology report in standardized clinical format... suitable for training a clinical NLP system.'' Simulates dataset curation for NLP training.
    \item \textbf{Teaching Case}: ``Rewrite as a clean, well-structured example report that could be used in a training dataset.'' Simulates educational dataset construction.
\end{itemize}

Temperature was set to 0.3 (with model-internal ``thinking'' disabled); each report was processed once per prompt, yielding 1,350 synthetic reports. We used Gemini 2.5 Flash~\cite{gemini2025} throughout; full prompt text is reproduced in the released code.

\subsection{Information Erosion Metrics}

\textbf{Entity erosion.} We extract medical named entities using scispaCy~\cite{neumann2019scispacy} with the \texttt{en\_core\_sci\_sm} model. For each pair:
\begin{equation}
    E_{\text{erosion}} = 1 - \frac{|\{e \in E_{\text{orig}} : \exists\, e' \in E_{\text{synth}},\; \text{match}(e, e')\}|}{|E_{\text{orig}}|}
\end{equation}
where $\text{match}(e, e')$ is true if $e$ and $e'$ are identical or one is a substring of the other. The substring criterion biases the metric conservatively. The denominator $|E_{\text{orig}}|$ is always positive in our sample: every original report contained at least three extracted entities, so the metric is well defined throughout. We additionally perform structured extraction via Gemini (temperature 0), identifying findings, negated findings, measurements, and recommendations in JSON.

\textbf{Hedging collapse.} We detect uncertainty markers via 18 regular-expression patterns matching: \emph{possible/possibly}; \emph{probable/probably}; \emph{suggest(s/ing/ive)}; \emph{cannot exclude/cannot rule out}; \emph{may represent/indicate/be}; \emph{could represent/indicate/be}; \emph{suspicious}; \emph{question(able)}; \emph{versus/vs}; \emph{differential includes}; \emph{concern(ing) for}; \emph{consider}; \emph{subtle}; \emph{equivocal}; \emph{uncertain}; \emph{indeterminate}; \emph{cannot be excluded}; and \emph{not entirely excluded/certain}. Hedging collapse is:
\begin{equation}
    H_{\text{collapse}} = 1 - \min(h_{\text{synth}} / h_{\text{orig}},\, 1)
\end{equation}
where $h_{\text{orig}}$ and $h_{\text{synth}}$ are the counts of detected hedging markers in the original and synthetic report, respectively. The metric is computed only for reports with $h_{\text{orig}} > 0$ ($n = 94$).

\textbf{Cross-modal alignment.} We embed each chest X-ray image and each report with the BiomedCLIP~\cite{zhang2023biomedclip} image and text encoders, respectively, and compute the cosine similarity between the (L2-normalized) image and text embeddings:
\begin{equation}
    \Delta_{\text{align}} = \text{sim}(I, T_{\text{orig}}) - \text{sim}(I, T_{\text{synth}})
\end{equation}
Positive $\Delta_{\text{align}}$ indicates the synthetic text is less aligned with the image. All 450 reports were matched to their paired chest X-ray images. We additionally report \emph{text-to-text similarity}, the cosine similarity between the BiomedCLIP text embeddings of the original and each synthetic report, as a content-overlap measure independent of the image. Report text was truncated to BiomedCLIP's 256-token limit (the first 1{,}000 characters) before encoding.

\subsection{Statistical Analysis}
We compare rare versus common pathology groups using Mann-Whitney $U$ tests. Our pre-specified hypothesis was directional (rare $>$ common); however, because the observed effects ran in the opposite direction, we report \emph{two-sided} tests as the primary analysis and treat directional results as exploratory. Across the nine comparisons (three metrics $\times$ three contamination types) we apply a Bonferroni correction, giving a family-wise threshold of $\alpha = 0.05/9 \approx 0.0056$. Cohen's $d$ is reported as the effect size, with positive $d$ indicating greater degradation in the rare group.

\section{Results}\label{sec:results}

\subsection{The Slop Paradox: Contamination Type Determines Degradation}
\begin{figure}[t]
	\centering
	\includegraphics[width=\textwidth]{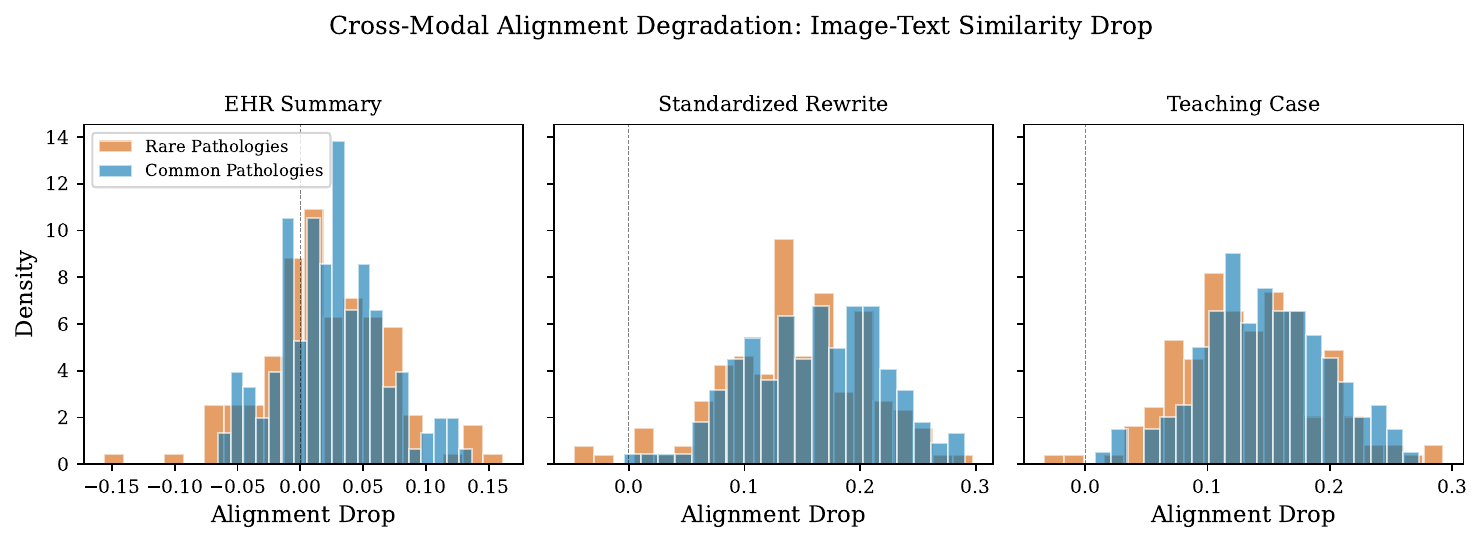}
	\caption{Distribution of cross-modal alignment drop (original minus synthetic image-text similarity) across three contamination types. Positive values indicate the synthetic text is less aligned with the image. EHR summaries cause minimal drift; standardized rewrites and teaching cases cause substantial misalignment.}\label{fig:alignment}
\end{figure}

We begin with the most striking finding: the contamination type that LLMs are most often used to produce, in the name of dataset improvement, is precisely the contamination type that causes the most severe cross-modal misalignment (Fig.~\ref{fig:alignment}). EHR summaries caused a mean alignment drop of 0.025 (SD = 0.043), while standardized rewrites caused 0.165 (SD = 0.064) and teaching cases 0.149 (SD = 0.053): a six-to-seven-fold difference. Original mean image-text similarity was 0.361. After EHR summarization, it dropped marginally to 0.336 (6.8\% relative). After standardized rewriting, it dropped to 0.196 (45.7\% relative), and after teaching case reformulation to 0.212 (41.3\% relative). Text-to-text similarity between original and synthetic followed a complementary pattern (EHR: 0.832; Std. Rewrite: 0.618; Teaching: 0.626), confirming that the two dataset-curation tasks restructure reports far more aggressively than EHR summaries.

\subsection{Overall Information Erosion}
Synthetic rewriting causes substantial information loss across all conditions. For EHR summarization, mean entity erosion was 51.4\% (SD = 0.217, $n = 450$): more than half of medical entities present in original reports were absent from synthetic summaries. Hedging collapse averaged 43.7\% (SD = 0.446, $n = 94$). Reports were compressed to 68.0\% of original length on average. Standardized rewrites and teaching cases showed lower entity erosion (26.8\% and 29.3\%) but expanded reports substantially (4.6$\times$ and 3.7$\times$ original length). Despite this expansion, hedging language was reduced by 32.3\% and 26.4\% respectively. Table~\ref{tab:summary} reports stratified statistics for the EHR Summary condition; Table~\ref{tab:summary_other} reports the corresponding statistics for the standardized rewrite and teaching case conditions.

\begin{table}[t]
\centering
\caption{Summary statistics for the EHR Summary contamination, by pathology group. Entity erosion and hedging collapse are proportions; compression ratio $<1$ indicates shorter synthetic text.}\label{tab:summary}
\footnotesize
\begin{tabular}{llccccc}
\hline
\textbf{Metric} & \textbf{Group} & \textbf{N} & \textbf{Mean} & \textbf{SD} & \textbf{Median} & \textbf{Range} \\
\hline
Entity Erosion   & Rare    & 150 & 0.481 & 0.199 & 0.462 & 0.08--0.92 \\
                 & Common  & 150 & 0.522 & 0.213 & 0.530 & 0.09--1.00 \\
                 & Normal  & 150 & 0.541 & 0.237 & 0.500 & 0.00--1.00 \\
Hedging Collapse & Rare    & 53  & 0.429 & 0.442 & 0.333 & 0.00--1.00 \\
                 & Common  & 30  & 0.367 & 0.440 & 0.000 & 0.00--1.00 \\
                 & Normal  & 11  & 0.667 & 0.447 & 1.000 & 0.00--1.00 \\
Compression      & Rare    & 150 & 0.673 & 0.281 & 0.649 & 0.13--2.15 \\
                 & Common  & 150 & 0.665 & 0.247 & 0.645 & 0.24--1.63 \\
                 & Normal  & 150 & 0.702 & 0.246 & 0.732 & 0.16--1.31 \\
\hline
\end{tabular}
\end{table}

\begin{table}[t]
\centering
\caption{Summary statistics for the Standardized Rewrite and Teaching Case contaminations, by pathology group. Entity erosion and hedging collapse are proportions; the length ratio is synthetic/original word count ($>1$ indicates the synthetic text is longer than the original). Hedging collapse is computed only over reports with hedging in the original.}\label{tab:summary_other}
\footnotesize
\begin{tabular}{lllccccc}
\hline
\textbf{Condition} & \textbf{Metric} & \textbf{Group} & \textbf{N} & \textbf{Mean} & \textbf{SD} & \textbf{Median} & \textbf{Range} \\
\hline
Std. Rewrite & Entity Erosion   & Rare   & 150 & 0.231 & 0.136 & 0.239 & 0.00--0.62 \\
             &                  & Common & 150 & 0.256 & 0.156 & 0.250 & 0.00--0.73 \\
             &                  & Normal & 150 & 0.318 & 0.219 & 0.297 & 0.00--1.00 \\
             & Hedging Collapse & Rare   & 53  & 0.322 & 0.438 & 0.000 & 0.00--1.00 \\
             &                  & Common & 30  & 0.332 & 0.457 & 0.000 & 0.00--1.00 \\
             &                  & Normal & 11  & 0.303 & 0.458 & 0.000 & 0.00--1.00 \\
             & Length Ratio     & Rare   & 150 & 3.905 & 3.675 & 3.059 & 1.00--33.00 \\
             &                  & Common & 150 & 4.164 & 3.330 & 3.482 & 1.39--29.00 \\
             &                  & Normal & 150 & 5.808 & 3.754 & 5.088 & 1.64--36.92 \\
\hline
Teaching     & Entity Erosion   & Rare   & 150 & 0.239 & 0.130 & 0.233 & 0.00--0.67 \\
             &                  & Common & 150 & 0.289 & 0.159 & 0.286 & 0.00--0.86 \\
             &                  & Normal & 150 & 0.350 & 0.199 & 0.333 & 0.00--0.89 \\
             & Hedging Collapse & Rare   & 53  & 0.273 & 0.420 & 0.000 & 0.00--1.00 \\
             &                  & Common & 30  & 0.267 & 0.412 & 0.000 & 0.00--1.00 \\
             &                  & Normal & 11  & 0.212 & 0.402 & 0.000 & 0.00--1.00 \\
             & Length Ratio     & Rare   & 150 & 3.560 & 2.382 & 2.786 & 1.45--16.87 \\
             &                  & Common & 150 & 3.360 & 2.166 & 2.732 & 0.84--13.27 \\
             &                  & Normal & 150 & 4.261 & 3.319 & 3.345 & 1.59--25.33 \\
\hline
\end{tabular}
\end{table}

\subsection{Entity Erosion, Hedging Collapse, and Compression}
\begin{figure}[t]
    \centering
    \includegraphics[width=\textwidth]{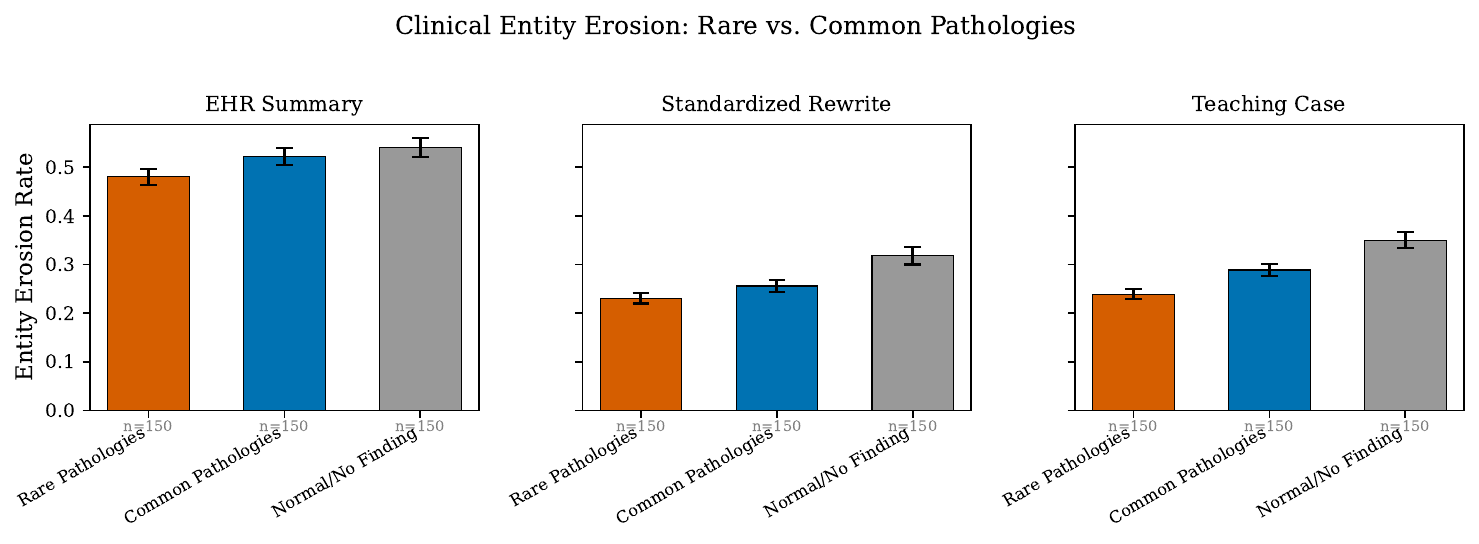}
    \caption{Entity erosion by pathology group and contamination type. Erosion shows no preferential effect on rare conditions; if anything, the common and normal groups erode slightly more. EHR summarization causes the highest erosion due to aggressive compression.}\label{fig:erosion}
\end{figure}

\begin{figure}[t]
    \centering
    \includegraphics[width=\textwidth]{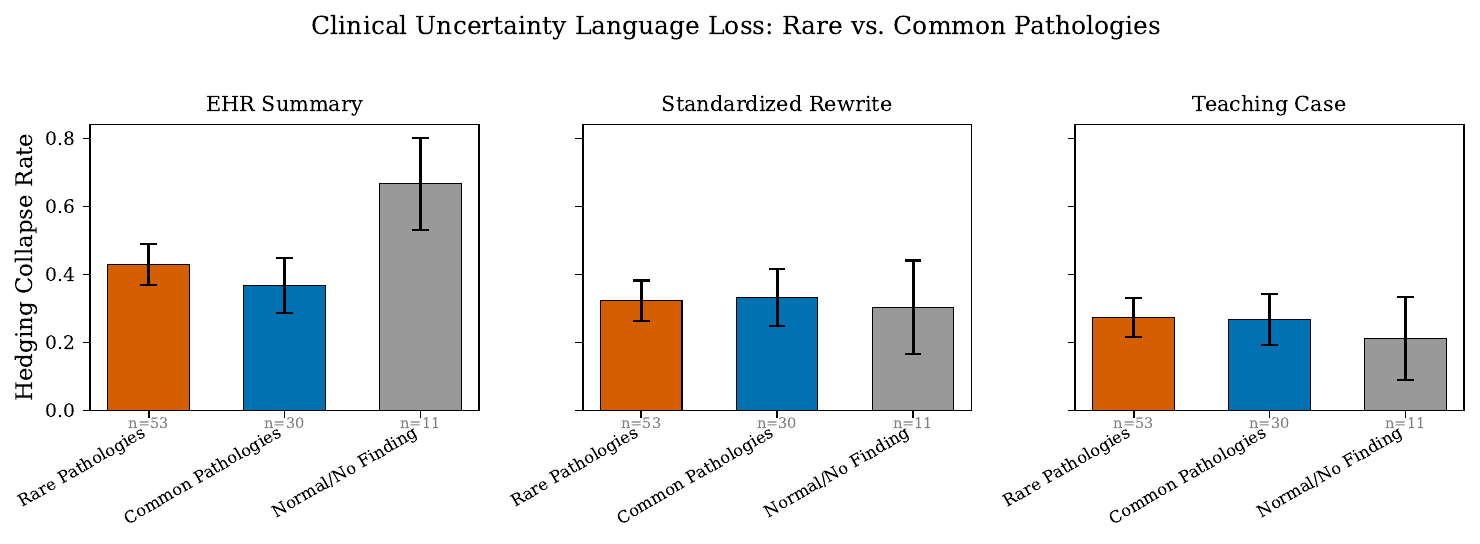}
    \caption{Hedging collapse by pathology group and contamination type. Only reports with hedging in the original are included ($n = 94$). Normal reports show the highest collapse rate.}\label{fig:hedging}
\end{figure}

Fig.~\ref{fig:erosion} shows entity erosion across pathology groups. EHR summarization produces the highest erosion ($\sim$50\%) due to aggressive compression. Standardized rewrites and teaching cases lose 27--29\% of entities despite producing longer texts: the added text consists of formatting and explanatory language rather than preserved clinical details. Hedging collapse (Fig.~\ref{fig:hedging}) is consistent across contamination types at 26--44\%. Normal/no-finding reports show the highest hedging collapse (66.7\% for EHR summaries), as the few hedging instances in these reports (e.g., ``no \emph{definite} pneumothorax'') are disproportionately lost when the entire report compresses to ``Normal chest X-ray.'' This finding has implications for documentation of pertinent negatives in clinical practice.

\begin{figure}[t]
    \centering
    \includegraphics[width=0.65\textwidth]{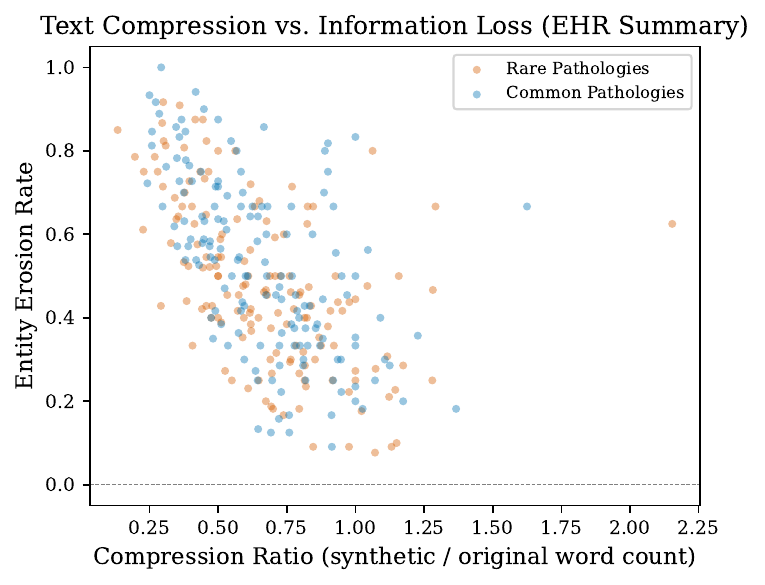}
    \caption{Compression ratio (synthetic/original length) vs. entity erosion for EHR summaries. Lower ratios, indicating more aggressive shortening, are associated with higher entity erosion; the trend holds across pathology groups.}\label{fig:scatter}
\end{figure}

Fig.~\ref{fig:scatter} shows the relationship between compression and erosion for EHR summaries. Reports compressed to 30--40\% of original length typically lose 60--80\% of entities. The relationship is nonlinear: moderate compression causes disproportionate entity loss, suggesting that entities removed first are not redundant but rather the specific details that distinguish one case from another.

\subsection{No Robust Differential Effect by Pathology Rarity}
We pre-specified the hypothesis that rare pathologies would suffer disproportionate information loss, motivated by distributional narrowing in model collapse research~\cite{ott2026context,anon_slop2025}. This directional hypothesis was not supported: in every comparison the rare group was degraded no more than the common group, and in most the direction was \emph{opposite} to prediction (e.g., entity erosion of 0.522 common vs. 0.481 rare for EHR Summary, $d = -0.199$). Treating the comparisons two-sided (Table~\ref{tab:tests}), three of the nine reached nominal significance, all in the opposite direction (common $>$ rare): teaching-case entity erosion ($p = 0.006$), standardized-rewrite alignment drop ($p = 0.032$), and teaching-case alignment drop ($p = 0.030$). After Bonferroni correction for the nine tests ($\alpha = 0.0056$), none of these remained significant. We therefore find no robust evidence that degradation depends on condition rarity, and what descriptive trend exists points toward marginally greater degradation of common conditions rather than rare ones. The practical consequence is unchanged: because degradation does not track rarity, contamination cannot be detected by monitoring model performance on specific condition classes.

\begin{table}[t]
	\centering
	\caption{Statistical comparisons between rare and common pathology groups (Mann-Whitney $U$ test, two-sided). Reported $p$-values are uncorrected; the Bonferroni threshold for the nine comparisons is $\alpha = 0.0056$, which no comparison meets. Positive $d$ indicates greater degradation in the rare group; all significant or near-significant effects are negative (common $>$ rare).}\label{tab:tests}
	\footnotesize
	\begin{tabular}{llcccccc}
		\hline
		\textbf{Condition} & \textbf{Metric} & \textbf{Rare} & \textbf{Common} & \textbf{$\Delta$} & \textbf{$d$} & \textbf{$U$} & \textbf{$p$} \\
		\hline
		EHR Sum.   & Entity Eros. & .481 & .522 & $-.041$ & $-.199$ & 10036 & .106 \\
		EHR Sum.   & Hedg. Coll.  & .429 & .367 & .062    & .141    & 857   & .530 \\
		EHR Sum.   & Align. Drop  & .020 & .025 & $-.005$ & $-.118$ & 10536 & .342 \\
		Std. Rewr. & Entity Eros. & .231 & .256 & $-.025$ & $-.170$ & 10430.5 & .275 \\
		Std. Rewr. & Hedg. Coll.  & .322 & .332 & $-.009$ & $-.021$ & 787   & .935 \\
		Std. Rewr. & Align. Drop  & .144 & .160 & $-.016$ & $-.272$ & 9643  & .032 \\
		Teach.     & Entity Eros. & .239 & .289 & $-.049$ & $-.341$ & 9205.5 & .006 \\
		Teach.     & Hedg. Coll.  & .273 & .267 & .006    & .014    & 802   & .941 \\
		Teach.     & Align. Drop  & .135 & .147 & $-.013$ & $-.235$ & 9622  & .030 \\
		\hline
	\end{tabular}
\end{table}

\section{Discussion}\label{sec:discussion}

\textbf{The slop paradox and its implications for multimodal training.} Our finding that contamination type, not clinical content, determines degradation magnitude has direct consequences. The two rewriting tasks most commonly associated with training dataset curation (standardization and teaching case preparation) produce the largest cross-modal alignment drops (16.5\% and 14.9\%). This creates a paradox: the very processes used to ``clean'' medical datasets for model training are the ones most likely to degrade the image-text correspondence multimodal models depend on. EHR summarization causes minimal alignment drift (2.5\%) despite high entity erosion (51.4\%), because it retains the core diagnostic conclusion while stripping detail, preserving broad semantic alignment. Standardized rewrites and teaching cases restructure the report entirely, introducing formatting and explanatory language that shifts the text embedding away from the visual content. This suggests practical risk stratification: AI-assisted documentation for point-of-care use poses relatively low risk to multimodal training, while LLM-based standardization for dataset construction introduces substantial alignment degradation.

\textbf{Clinical significance of hedging collapse.} The 26--44\% destruction of hedging language across contamination types represents a patient safety concern independent of training data contamination. When a report states ``cannot exclude early pneumonia'' and the AI summary renders this as ``no acute findings,'' the clinical uncertainty that should trigger follow-up is silently eliminated. Rabaey et al.~\cite{rabaey2025modeling} demonstrated that hedging phrases convey specific probability levels, and Goodman et al.~\cite{goodman2024ai} documented ``complete-the-narrative'' errors where LLMs convert uncertain pictures into definitive ones. Our results quantify this at scale. The highest collapse rates appear in normal reports (66.7\%), where structured negative findings (``no definite consolidation,'' ``lungs are clear bilaterally without focal opacity'') are compressed to generic statements. In clinical practice, the documentation of pertinent negatives establishes the radiologist's scope of review and provides a baseline for future comparisons.

\textbf{Why the rare/common hypothesis failed.} Our pre-specified directional hypothesis was not supported, and the descriptive trend ran opposite to prediction. Two-sided tests showed nominally higher degradation in \emph{common} pathologies for three of nine comparisons (teaching-case entity erosion $p = 0.006$, standardized-rewrite alignment drop $p = 0.032$, teaching-case alignment drop $p = 0.030$), though none survived Bonferroni correction ($\alpha = 0.0056$). Two factors likely explain the absence of a rare-pathology penalty. First, our experiment measures single-step compression, not recursive distributional bias: the LLM is summarizing individual reports, where compression dynamics are similar regardless of pathology frequency. The distributional narrowing that drives model collapse~\cite{ott2026context} operates across training iterations, not within a single generation step. Second, the IU dataset contains short reports (mean 41 words); with limited text, rare and common conditions are described with similar brevity, so the LLM compresses them similarly. A plausible account of the slight common-pathology lean is that LLMs aggressively normalize high-frequency, familiar clinical phrases into generic templates, whereas rarer entities, being more distinctive tokens, occasionally resist this smoothing; because the effect does not survive correction, the operative conclusion is that condition rarity does not reliably predict degradation. We report this result because the hypothesis was pre-specified and the test was properly powered, and because the absence of a rarity signal is itself informative: monitoring model performance on specific condition classes will not reveal contamination.

\textbf{Limitations.} Our experiment measures single-step erosion, not recursive model collapse. We demonstrate \emph{preconditions} for training data degradation but do not demonstrate downstream performance degradation in a trained model; future work should conduct retraining experiments on contaminated versus clean datasets. The IU dataset is relatively small with short reports; results may not generalize to longer clinical narratives. The hedging analysis was limited to 94 reports containing uncertainty language, reducing statistical power. Our entity erosion metric uses substring matching, which biases conservatively toward underestimating erosion. BiomedCLIP's 256-token text limit truncates longer synthetic reports (particularly standardized rewrites at 4.6$\times$ original length), which may partially explain alignment drops for these conditions. Even if truncation accounts for some of the standardized rewrite drop, however, it cannot explain the comparable drop observed for teaching cases (3.7$\times$ original length, also exceeding the token limit), so the underlying pattern that dataset-curation tasks drift further from image content than EHR summaries is robust to this concern. Finally, our pathology grouping is imperfect: rare labels took precedence in assignment, so the rare group contains reports that also carry common pathologies, and the normal group includes reports whose only findings fall in the 2--5\% mid-frequency band rather than being strictly normal. Both choices blur the rare-versus-common boundary and would attenuate any genuine rarity effect, so the absence of a robust differential should be read as ``not detectable under this stratification'' rather than as proof of strict equivalence.

\textbf{Governance implications.} Three recommendations follow. First, institutions deploying AI-assisted documentation should implement entity preservation audits comparing the clinical entity content of AI-generated summaries against source reports before summaries enter permanent records. Second, organizations constructing multimodal training datasets should avoid LLM-based text standardization or reformulation as a preprocessing step without measuring alignment degradation. Third, data provenance tracking should extend to clinical text: when a radiology report is AI-summarized, this transformation should be recorded in metadata so downstream consumers can distinguish original from synthetic text. These recommendations align with emerging frameworks for AI data governance in healthcare~\cite{bailo2026governing} and broader calls for synthetic content provenance~\cite{anon_slop2025}.

\section{Conclusion}\label{sec:conclusion}

We presented a controlled measurement of information degradation in AI-generated radiology reports across three clinically realistic contamination vectors. Our results demonstrate that synthetic rewriting causes substantial, systematic information loss: 51.4\% of clinical entities are eroded in EHR summaries, 26--44\% of hedging language is destroyed across all rewriting types, and cross-modal image-text alignment drops by up to 16.5\% for dataset-curation-oriented tasks. This degradation showed no preferential effect by pathology frequency, with no rare-versus-common difference surviving multiple-comparison correction, making it invisible to condition-specific performance monitoring. The dominant determinant of degradation magnitude is the type of AI rewriting task, not the clinical content: the standardized rewrites and teaching case reformulations most commonly used for training dataset curation cause six to seven times as much cross-modal misalignment as EHR summarization.

These findings have immediate practical implications. AI-assisted clinical documentation should be subject to entity preservation auditing. Multimodal training dataset construction should not rely on LLM-based text standardization without measuring alignment impact. And data provenance mechanisms should distinguish original clinical text from AI-generated derivatives throughout the medical data pipeline.

\bibliographystyle{unsrt}
\bibliography{references}

\end{document}